\RequirePackage{snapshot}
\documentclass[10pt,twocolumn,letterpaper]{article}
\newif\ifrev

\usepackage{times}
\usepackage{epsfig}
\usepackage{graphicx}
\usepackage{amsmath}
\usepackage{amssymb}

\usepackage{xspace}

\def\abstract{\centerline{\large\bf Abstract}\vspace*{12pt}\it}

\usepackage{xspace}
\makeatletter
\DeclareRobustCommand\onedot{\futurelet\@let@token\@onedot}
\def\@onedot{\ifx\@let@token.\else.\null\fi\xspace}
\def\eg{\emph{e.g}\onedot}

\def\etal{\emph{et al}\onedot}
\makeatother

\usepackage{titlesec}
\usepackage{etoolbox}
\makeatletter
\patchcmd{\ttlh@hang}{\parindent\z@}{\parindent\z@\leavevmode}{}{}
\patchcmd{\ttlh@hang}{\noindent}{}{}{}
\makeatother

\titlespacing {\section}
{0pt}{1.2ex plus 1ex minus .2ex}{1.0ex plus .2ex}
\titlespacing {\subsection}
{0pt}{1.0ex plus 1ex minus .2ex}{1.0ex plus .2ex}
\titleformat{\section}
  {\normalfont\fontsize{12}{15}\bfseries}{\thesection.}{0.3em}{}
\titleformat{\subsection}
  {\normalfont\fontsize{11}{15}\bfseries}{\thesubsection}{0.3em}{}

\usepackage{caption}
\usepackage{gensymb} 
\usepackage{textcomp} 
\usepackage{enumitem}
\setitemize{noitemsep,topsep=0pt,parsep=0pt,partopsep=0pt}

\ifrev
  \newcommand{\AF}[1]{{\color{blue} [AF: #1]}}
  \newcommand{\OF}[1]{{\color{green} [OF: #1]}}
  \newcommand{\OD}[1]{{\color{violet} [OD: #1]}}
\else
  \newcommand{\AF}[1]{}
  \newcommand{\OF}[1]{}
  \newcommand{\OD}[1]{}
\fi

\usepackage[pagebackref=true,breaklinks=true,letterpaper=true,colorlinks,bookmarks=true]{hyperref}

\begin{document}

\title{See far with TPNET: a Tile Processor and a CNN
Symbiosis}
\date{} 
\author{
Andrey Filippov \qquad Oleg Dzhimiev \\
Elphel, Inc.\\
1455 W. 2200 S. \#205, Salt Lake City, Utah 84119 USA \\
{\tt\small \{andrey,oleg\}@elphel.com}
}

\maketitle
\thispagestyle{empty}

\begin{abstract}
\label{sec:abstract}
Throughout the evolution of the neural networks more specialized cells were
added to the set of basic building blocks. These cells aim to improve training
convergence, increase the overall performance, and reduce the number of required
labels, all while preserving the expressive power of the universal network.
Inspired by the partitioning of the human visual perception system between the eyes and the
cerebral cortex, we present TPNET, which offloads universal and
application-specific CNN from the bulk processing  of the high resolution pixel
data and performs the translation-variant image correction while delegating all
non-linear decision making to the network.
 
In this work, we explore application of TPNET to 3D perception with a
narrow-baseline (0.0001-0.0025) quad stereo camera and prove that a trained
network provides a disparity prediction from the 2D phase correlation output by
the Tile Processor (TP) that is twice as accurate as the prediction from a
carefully hand-crafted algorithm. The TP in turn reduces the dimensions of the
input features of the network and provides instrument-invariant and
translation-invariant data, making real-time high resolution stereo 3D
perception feasible and easing the requirement to have a complete end-to-end
network.
\end{abstract}

\section{Introduction}
\label{sec:introduction}
 
We consider our work to contribute the following:
\begin{itemize}
  \item State-of-the-art narrow-baseline stereo camera
  which provides robust 3D measurements with 0.05 pixel accuracy and which is capable of operating at distances of hundreds
  to thousands meters (farther than automotive LIDAR and ToF
  cameras); and
  \item TPNET as a framework for partitioning the larger network into
  sensor-variant (``eyes'') and sensor-invariant (``brain'') subsystems, without
  sacrificing any of the capabilities of true end-to-end networks.
\end{itemize}

The recent development of vision perception systems is defined not only by
advances in new network architectures but also by the emergence and general
availability of new sensor technologies.
The appearance of direct distance measurements with \mbox{LIDAR} and ToF
cameras, which provide excellent range precision but low image plane resolution,
triggered the development of the fusion of such sensor data with high resolution
conventional images (Yang~\etal~\cite{yang2007spatial},
Park~\etal~\cite{park2011high}, Gu~\etal~\cite{gu2017learning}).

The cellphone camera revolution contributed to the development of the
Structure-from-Motion (SfM) 3D scene reconstruction. Recently,
Torii~\etal~\cite{torii2018structure} suggested a two-stage network that first
extracts features with VGG16 ($300\times150\times256$ tensors from $1600\times1200$ images),
then matches low-resoluton features to establish initial
correspondences and improves the keypoints localization to a single-pixel
resolution and builds 3D model from the N-best images. Other technological advances
caused by the widespread adoption of cellphone cameras include methods for the enhancement of their images, such as
motion blur elimination (Zhang~\etal~\cite{zhang132017learning}) and Electronic
Rolling Shutter (ERS) distortion correction with egomotion estimation from the
video frames sequence (Forss{\'e}n and Ringaby~\cite{forssen2010rectifying}) or
even from a single frame by automatic feature extraction of four
straight in real-world lines (Lao and Ait-Aider~\cite{lao2018robust}).

The diversity of research in the field of vision-based 3D perception is
handicapped by the limitations of publicly available datasets:
Pingerra~\etal~\cite{pinggera2014know} noticed that popular Middlebury and KITTI
frameworks do not sufficiently treat local sub-pixel matching accuracy. We would
add that such datasets are based on conventional binocular stereo
and do not provide the data needed for deep subpixel calibration,
effectively removing large application classes from consideration by machine
learning researchers.

Most ML-based vision perception systems, even nominally end-to-end ones, start
from the RGB images, usually rectified for compatibility with the
translation-invariance nature of the CNN. The raw sensor data is neither
rectified, no RGB, but rather a Bayer mosaic, normally consisting of a
repeating $2\times2$ (RG/GB) pixels pattern. Khamis~\etal~\cite{khamis2018stereonet}
achieved 1/30th of a pixel precision with an end-to-end network, that used
\textit{synthetic} data, same approach with KITTY~2015~\cite{Menze2015CVPR} did
not result in such precision. Bayer-to-RGB conversion is the most
lossy part of the camera image capturing, and advanced networks like the one
developed by Chen~\etal~\cite{chen2018learning} for low-light imaging bypass
color conversion and use raw Bayer pixel data instead.

This leads to conflicting system requirements: on the one hand, the low-level
image processing (color conversion, rectification) leads to loss of the
important sensor data; on the other hand, resorting to the raw sensor data makes
the whole system hardware-dependent and complicates the knowledge transfer and
inference of the trained network.

We address these challenges by proposing a ``network-friendly'' system that adjusts the hardware, low-level
processing and universal subnet to match existing DNN solutions. We
developed a complete prototype system that outputs X3D (Brutzman and
Daly~\cite{brutzman2010x3d}) scene models, but this is beyond the scope of this
work. We focus here on the most under-explored part of the 3D vision perception
that can be combined with and incorporated into other ML systems.

\section{Related work}
\label{sec:related}
\subsection{Long range stereo vision}
\label{subsec:long-range-stereo}

Binocular stereo cameras for distance measurement and 3D scene reconstruction
were very popular some 20 years ago and in 2002 Scharstein and Szeliski
presented~\cite{scharstein2002taxonomy} taxonomy of dense two-frame stereo
correspondence algorithms, at that time it did not distinguish between
traditional and ML-based approaches. Since then these applications
gradually lost their popularity for several reasons:
\begin {itemize}
  \item  direct active distance measurements with LIDAR and ToF cameras provide
  higher accuracy in most cases
  \item phone camera revolution made it easy to capture multiple views of
  the 3D objects to apply SfM processing.
\end{itemize}
There are still application areas where passive vision-based systems are
preferable -- in addition to obvious military ones where advertising yourself with the lasers
is not acceptable, passive systems may have advantage for the longer range than
practical for the automotive LIDAR scanners (over 200~meters) and for low power
systems where illuminating environment with your own photons may be costly.

These considerations focus our work on the long range (\textgreater100~m),
narrow-baseline cameras. For these cameras (Bayer mosaic
color $2592\times1936$, 2.2$\mu m$ pixels,
FoV=$60\degree(h)\times45\degree(v)$, baseline=258~mm) 100~m range
corresponds to 5~pixel disparity, so the main challenge is to achieve accurate
subpixel resolution that depends on:
\begin {itemize}
  \item optical-mechanical stability of the system
  \item pixel noise
  \item image processing methods
\end{itemize}

Pingerra~\etal~\cite{pinggera2014know} provided thorough comparison of multiple
fractional pixel calculation methods and noticed that while results were varying
by less than 1/30~pixel across all algorithms, the mean disparity error caused
by a
deviation in a
camera calibration reached 1/10~of a pixel.
We use thermally compensated sensor front ends with non-adjustable lenses
relying on DoF of the small pixel cameras.

Nature of the various 3D capturing tasks involves processing of different number
of pixels, influencing the disparity accuracy. The objects of interest may be
small clusters of pixels, \eg a distant flying drone, an edge of the foreground
object or a textured (often poorly) surface. Small clusters case falls into the
well investigated area of Particle Image Velocimetry (PIV) prone to the
pixel-locking effect, described by Fincham and Spedding~\cite{fincham1997low},
Chen and Katz~\cite{chen2005elimination} proposed method of reducing this effect
to under 0.01~pix for clusters of $4\times4$~pixels and above,
Westerweel~\cite{westerweel2000effect} studied influence of the pixel geometry
on correlation resolution. Pixel locking for stereo disparity may
occur for larger patches: Shimizu and Okutomi~\cite{shimizu2005sub} measured
it by
moving the target and then proposed compensation. 
Sabater~\etal~\cite{sabater2011accurate} considered theoretical limits of the
disparity accuracy in the presence of pixel noise.

Phase correlation (PC) in the Frequency Domain (FD) is free of 
pixel locking: Hoge\cite{hoge2003subspace} used Singular Value
Decomposition (SVD) for disparity from PC
of the 
MRI images, Balci and Foroosh\cite{balci2005inferring}
developed plane fitting to the PC for satellite imagery.
Morgan~\etal~\cite{morgan2010precise} reported
0.022~pix RMSE for
$32\times32$
window model photos with fine texture using FD PC.
Both methods (SVD and phase plane fitting) best work with large
windows and small disparity variations,
but direct feeding of the small window PC data to the trainable network may
resolve ambiguity inherent to these methods.

We handle pixel locking in two ways: by fast converging iteration with lossless
pre-shifting of the patches in FD, and then feed the network with that
correlation data. Direct feeding of the FD representation to the network may be
beneficial as it tends to concentrate important information in a
small number of coefficients, but the pixel domain PC also has
the same property to keep relevant data compact, we will try to concatenate both
types of data.

Edges of the foreground objects are very important for stereo image matching as
they correspond to photometric discontinuities
and are the major contributors to the feature-based image matching, such as SIFT
(Lowe~\cite{lowe1999object}) and HOG descriptors used for human detection by
Dalal and Triggs~\cite{dalal2005histograms}.
In the case of 
dense correspondence
edges are handled by a separate term in SGM algorithm
(Hirschmuller~\cite{hirschmuller2005accurate}) and its derivatives. SGM was
developed for traditional implementation (including FPGA RTL) but it still
remains efficient when combined with the network by Zbontar and
LeCun~\cite{zbontar2015computing}.

\begin{figure}[t]
\begin{center}
\includegraphics[width=1.0\linewidth]{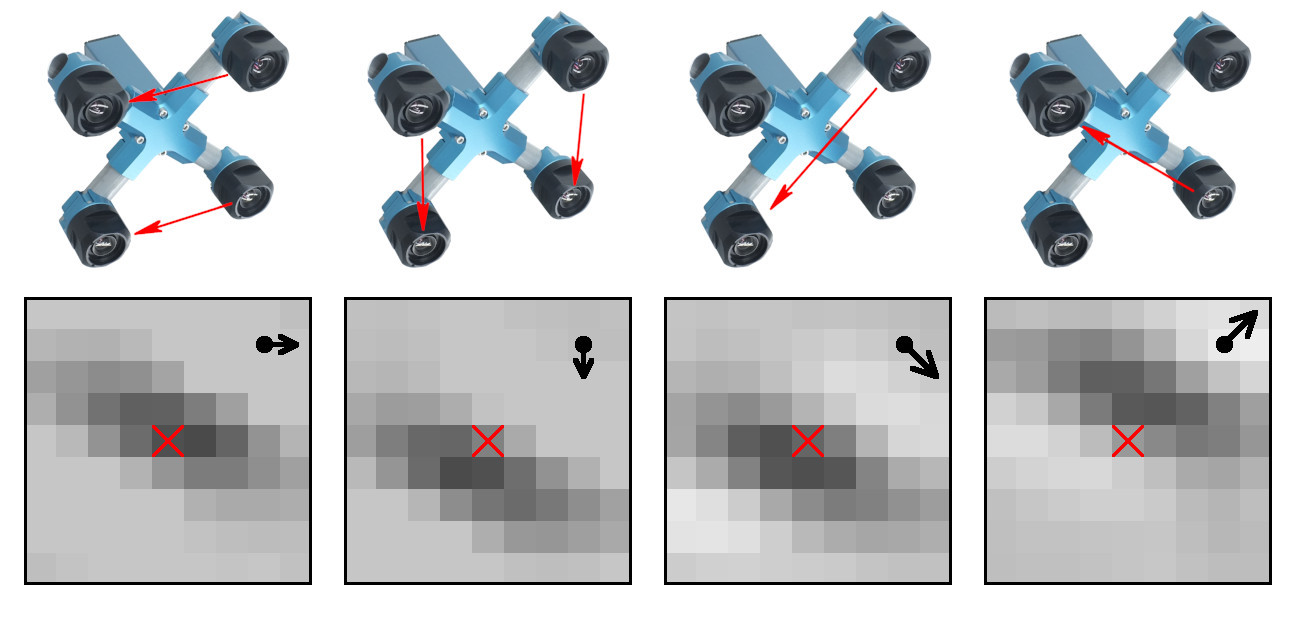}
\end{center}
\caption{Quad stereo camera and 2D phase correlation (PC) samples between
   different image pairs.}
\label{fig:cameras-correlations-4-dirs}
\label{fig:onecol}
\end{figure}

Regardless of the image processing method,
binocular stereo cameras are insensitive to the edges parallel to the epipolar
lines. Commonly used systems with the two horizontally offset cameras cannot
measure disparity of the horizontal linear features. We use quad camera system
and process four directions for 2D PC as shown in
Figure~\ref{fig:cameras-correlations-4-dirs}, these 45\textdegree orientations
provide almost omnidirectional representations of the edges. 

\subsection{FD and the neural networks}
\label{subsec:fd-neural}
FD operations attract ML researches, one reason is that in FD
computationally-intensive convolutions become trivial pointwise multiplications.
Another -- that the first CNN layers as visualized by deconvnet
(Zeiler~\etal~\cite{zeiler2010deconvolutional}) exhibit Gabor-like patterns
similar to FD representation and can potentially replace these layers with
optimized modules.

Mathieu~\etal~\cite{mathieu2013fast} developed new CUDA FFT implementation and
compared performance for variable image sizes (16-64) convolution with smaller
$7\times7$ kernels. Vasilache~\etal~\cite{vasilache2014fast} evaluated
GPU implementations with NVIDIA cuFFT and their fbfft, reporting
performance gains for the kernels above $3\times3$.
Brosch and Tam~\cite{brosch2015efficient} studied two-layer convDBN
trained in FD for 2D and 3D MRI images, achieving $200\times$ performance gain.
Chen~\etal~\cite{chen2016compressing} used Discrete Cosine Transform (DCT)
instead of DFT followed by hash function to reduce width of the filters.
Rippel and Adams~\cite{rippel2015spectral} found that use of the 2D FD input
to CNN followed by spectral pooling has advantages over max pooling in the
pixel domain data. Zhao~\etal~\cite{zhao2018sound} found that audio
spectrograms improve performance from raw waveforms.

\begin{figure}[t]
\begin{center}
\includegraphics[width=0.85\linewidth]{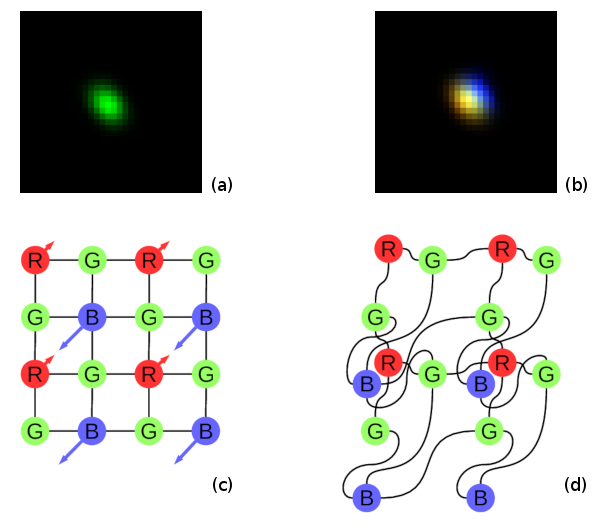}
\end{center}
\caption{Bayer mosaic and lateral chromatic aberration. (a)~Monochrome Point Spread Function (PSF).
(b) Composite color PSF. (c) Bayer mosaic of the sensor (direction of aberration shown).
(d) Distorted mosaic for chromatic aberration of (b).}
\label{fig:chromatic}
\end{figure}

In this work we use fixed-size ($16\times16$ stride 8) Modified Complex
Lapped Transform (MCLT) introduced by Princen~\etal~\cite{princen1987subband}
and later applied to 2D image compression by De Queiroz and
Tran~\cite{de2001lapped} (we describe its details in 
Section~\ref{subsec:mclt}).
The focus of this work is to introduce an efficient interface
between the low-level hardware-dependent image processing and
translation-invariant CNN rather than to optimize convolutions in the network
itself.

\section{TPNET implementation}
\label{sec:tpnet}
\subsection{Best of the raw and the hardware-invariant}
\label{subsec:raw-invariant}

Most dense image matching 3D perception systems -- both traditional and those
based on DNN -- try to determine each pixel's disparity, typically with
SGM~\cite{hirschmuller2005accurate} or its modifications.
We split this task into a separate low spacial resolution disparity measurement,
to be followed by fusion with the high-resolution RGBA images (where the alpha
channel is calculated from the photometric differences between the images),
similar to fusion for ToF cameras~\cite{gu2017learning, park2011high,
yang2007spatial}.

Replacing a single set of the raw images with multiple views of the same data
resolves the conflicting requirements of the instrument and
translation-invariance and the preservation of all of the relevant source data.
Generally, the correction and rectification of image optical aberrations
involves re-sampling that either requires up-sampling or adds quantization noise
that jeopardizes subpixel accuracy.
The correlation between the two tiles can be performed in the FD after
performing the lossless phase rotation equivalent to the fractional pixel shift
in the pixel domain. As the result, we can simultaneously obtain the RGBA data
with pixel resolution and per-tile 2D correlation data for disparity
measurements with full subpixel accuracy preserved. Both of these views provide
instrument and translation-invariant data compatible with CNN, but neither of
them can be derived from the other.

Similar to Chen~\etal~\cite{chen2018learning}, we bypass Bayer to RGB conversion
and use the raw data. The resolution of modern small pixel sensors is higher
than that of the lenses, and aberrations in the off-center areas may
even
exceed the pixel
pitch,
invalidating assumption that red pixel is always located in the middle between
four green ones (Figure~\ref{fig:chromatic}).
In this work, we process each color
channel separately, apply individual aberration corrections, and merge results
after the PC. Texture images merged from the four sensors using
predicted disparity have Bayer-related artifacts attenuated from the
single-camera artifacts, as the color patterns have random offsets. The final
improvement of the texture images can be done with DNN (together with denoising
and super-resolution), following the approaches of
Gharbi~\etal~\cite{gharbi2016deep}, Syu~\etal~\cite{syu2018learning} with
modifications that consider multi-camera images of the same patches.

\begin{figure*}
\begin{center}
\includegraphics[width=0.8\linewidth]{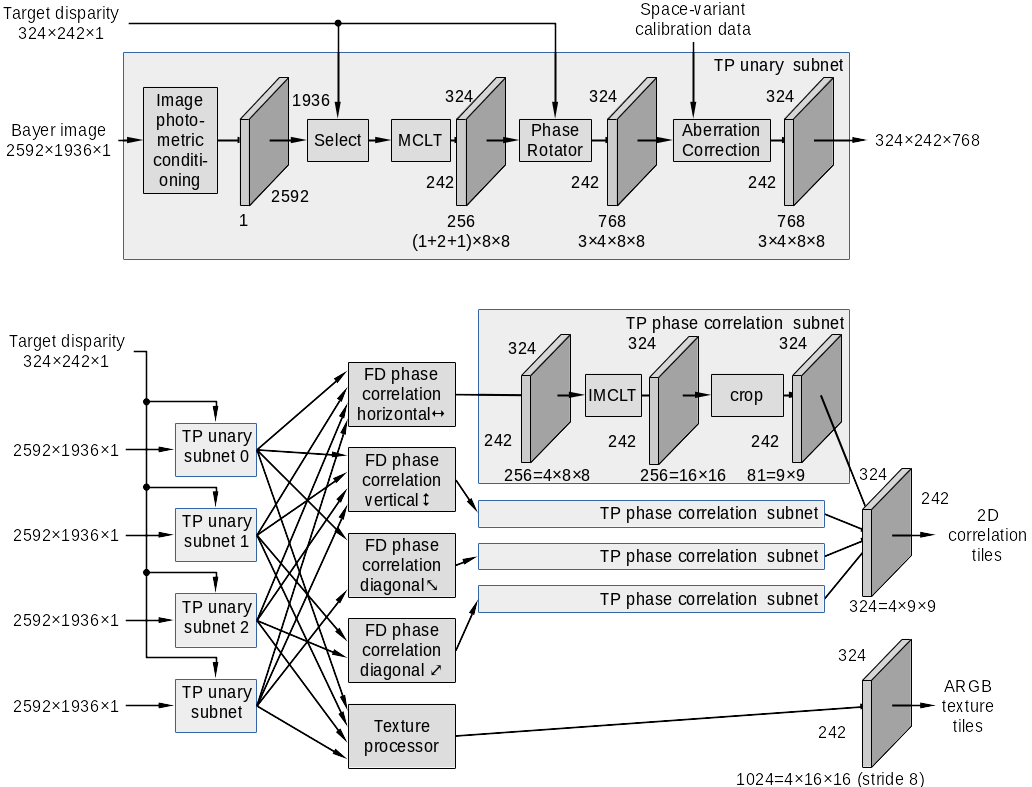}
\end{center}
   \caption{Tile Processor - a frequency domain pre-processor for the DNN}
\label{fig:tile_processor}
\end{figure*}

\subsection{More non-collinear cameras}
\label{subsec:more-cameras}

Use of the quad stereo camera instead of the conventional binocular layout
extends SGM~\cite{hirschmuller2005accurate} approach to the camera design -- we
are not just calculating the disparity cost along multiple directions, but are
rather measuring the disparity along multiple epipolar lines, allowing the
downstream network to increase the weights of the correlations for the pairs
orthogonal to the foreground object edge and ignore  correlations produced by
the background texture.

The increased  number of image sensors as compared to the traditional binocular
system is not expensive when it does not lead to an increase of the overall
dimensions. Extra images will be reused to enhance the resolution of the
combined image (Jeon~\etal~\cite{jeon2018enhancing}) and to improve the S/N
ratio of the
sensors. Using more sensors when
the parallax is compensated allows simultaneous HDR
(Popovic~\etal~\cite{popovic2016multi}) and multispectral 3D imaging
(Neukum~\etal~\cite{neukum2017high}).

\subsection{Frequency Domain (FD) processing} 
\label{subsec:fd}

The initial TPNET implementation includes Tile Processor (TP), shown in
Figure~\ref{fig:tile_processor} and DNN (Figure~\ref{fig:diag-net}) fed with the
2D correlation data from TP. TP performs the image conversion to the FD and the phase
rotations (subpixel shifts), calculates the convolutions, PC
and other FD operations,
then converts the result arrays back to the pixel domain. 
We use fixed-size tiles that are large enough for deep subpixel disparity
resolution; namely, tiles which are large enough to provide efficient pooling
and reduction of disparity space image dimensions, but small enough to avoid
scale and rotation mismatch.
They are also sufficiently small so as to reduce the number of different
disparity values sharing the same tile -- usually just one or two for the edge
of the foreground tile over the background.

\subsection{Modified Complex Lapped Transform for FD}
\label{subsec:mclt}

The tile receptive field is $16\times16$ pixels, and, if applied to the full image
with stride 8 (50\% overlap in each direction),
the operation is reversible due to the \textit{ perfect reconstruction property} of
the MCLT~\cite{princen1987subband,malvar1990lapped,de2001lapped}.
Tiles do not have to be processed for the whole image, and each of them may have
different
``target disparity''~(TD) that defines the selection of the image patches to
be matched and the subpixel fraction shift to be applied
before combining. The TD is analogous to eye convergence,
but it is applied individually to each tile, not to the whole images.

MCLT implementation is based on Discrete Cosine Transform type IV (DCT-IV) and
its sine counterpart DST-IV. MCLT is a generalization of MDCT - transform which
is sufficient for compression but does not have
the full convolution-multiplication property. For the single dimension, the
forward transform MDCT normally uses the half-sine window function for the input
data and ``folds'' $2N$ input sequence to an $N$-long ($N=8$ in our case) one.
The subsequent DCT-IV transforms eight input values into eight outputs. When
used for compression, these
coefficients are subject to quantization and transmission to the decoder.
The decoder performs an $N$-long inverse transform (in case of DCT-IV it is
equal to the direct one), unfolds it to a $2N$-long sequence, and multiples it
by the window function again (where the window satisfies the required
Princen-Bradley condition~\cite{princen1987subband}). The final step of the
decoding would be to add together the individual $2N$-long sequences with
overlap$N$, and without quantization the restored sequence will exactly match
the initial one, losing only the first and the last $N$ of continuous samples.

For the full convolution-multiplication property to be valid,
the complex-valued transform is needed,
and for the single dimension the MCLT consists of a pair of MDCT and MDST, based on
DCT-IV and DST-IV respectively, such that overlapping real-valued $2N$-long
sequences are converted to pairs of $N$-long ones. Similarly, two dimensional
$2N\times2N$ tiles are converted to four
 (one for each variant of horizontal and vertical DCT-IV and DST-IV)
$N\times N$ tiles.
In case of $8\times8$
transforms, the $16\times16$ overlapping tiles are converted to
$4\times8\times8$ real-valued tensors equivalent to twice larger $16\times16$
DFT arrays with the same number of elements, but with complex instead of real values.

We modified MCLT algorithm to correctly handle translation-variant tile
offsets by applying up to $\pm0.5$ pix shifts to the original half-sine
windows.
 
Our additional $3\times$ optimization is applied to the direct MCLT
conversion of the Bayer mosaic images. Monochrome transformation of $16\times16\times1$
into $4\times8\times8$ tensor requires four $8\times8$ DTT-IV (
``Trigonometric'',
DTT~=~D\{C,S\}T)
transforms. For the color
Bayer mosaic tile of the same $16\times16\times1$ size, producing $3\times4\times8\times8$
output requires
equal
number of DTT-IV operations (that is, four: 1~for
red, 1~for blue and 2~for green), instead of the 12 needed for full RGB
$16\times16\times3$ conversion.
 
\subsection{Full 2D correlation instead of 1D epipolar}
\label{subsec:2d-1d}

TP exploits the convolution-multiplication property for efficient
implementation of the convolution and correlation.
Rather than the conventional 1D correlation along the epipolar lines, we use
full 2D correlation. Computationally, it does not require additional resources,
as the 2D tiles are already available in FD after aberration correction, and the
2D correlation output for all pairs
(Figure~\ref{fig:cameras-correlations-4-dirs}) provides the network with
additional data about the edge direction; the relative importance of the pairs
for the disparity measurement may be obtained from  consolidating the data from
several neighboring tiles, allowing the network to follow the linear features
and to improve the S/N ratio in low textured areas.
The 2D disparity vector is also used to
correct
the misalignment of the cameras. In addition to correlations between the
tiles of the simultaneously captured images, the same TP can calculate motion
vectors and measure optical flow from the consecutive frames of the
camera.

\subsection{Tile Processor pipeline}
\label{subsec:tp-pipeline}
Incoming Bayer mosaic images (Figure~\ref{fig:tile_processor}) from each of the
four sensors are first processed by identical channels that output FD tiles
which preserve all the input data and thus may be converted back. In the current
implementation, all of the required calibration data (such as the space-variant
convolution kernels for aberration correction) are calculated with specially
designed software and calibration setup. All the transformations are linear and
differentiable with respect to both pixel values and calibration parameters, and
they are immanent to the camera hardware, not to the specific application.
This makes it possible to develop a trainable network to find the calibration
kernels and the subcameras' global intrinsic (such as focal length and radial
distortion) and extrinsic (relative pose) parameters which is independent of the
overall system application. Next, operations are performed for each tile
independently which are parallelized according to the hardware capabilities (we
developed RTL, CPU and GPU code, released as FLOSS). Tile coordinates are
defined for the virtual camera located in the middle between the physical
cameras. This virtual camera has radial distortion calculated as a best
simultaneous fit for all the four actual cameras. The remaining deviations of
each physical camera from the virtual one are considered aberrations and are
treated the same way as chromatic and other aberrations by specifying centered
deconvolution kernel and pixel offsets with fractional pixel resolution.

Each tile receives the target disparity value, uses extrinsic and distortion
parameters to find nearest convolution kernel for each color, reads in that
kernel, interpolates kernel center offset and calculates the full coordinates of
each color component tile center and reads the corresponding image data from the
$1936\times2592$ array into combined $16\times16$ buffer. The pixel window
selection accommodates the integer part of the total pixel offset, while the
fractional part is applied twice: first it modifies the nominally half-sine
window function for MCLT, then after MCLT it is applied as a phase rotation in
the FD, equivalent to the lossless fractional pixel shift. At this stage, the
data does not have a real-valued pixel domain representation and so has to be
processed in FD before the inverse transformation. It is point-wise multiplied
by the calibration kernel, resulting in a $3\times4\times8\times8$ tensor of
corrected FD data for each tile, concluding the unary processing.

FD image tensors $242\times324\times3\times4\times8\times8$ are used for 2D
PC and texture processing. The 2D PC in the FD
consists of pointwise complex multiplication, followed by weighed averaging
between color channels and normalization. Then, each tile is converted to the
pixel domain (using DCT-II/DST-II), and each $16\times16$ output is cropped to
the center $9\times9$ shown in Figure~\ref{fig:cameras-correlations-4-dirs}.

The texture processor uses the same FD representations of corrected and shifted
according to the specified disparity image tiles. They are combined and
inverse-transformed to the pixel domain with the alpha-channel obtained from the
pixel value differences between subcameras.

\begin{figure}[t]
\begin{center}
\includegraphics[width=0.95\linewidth]{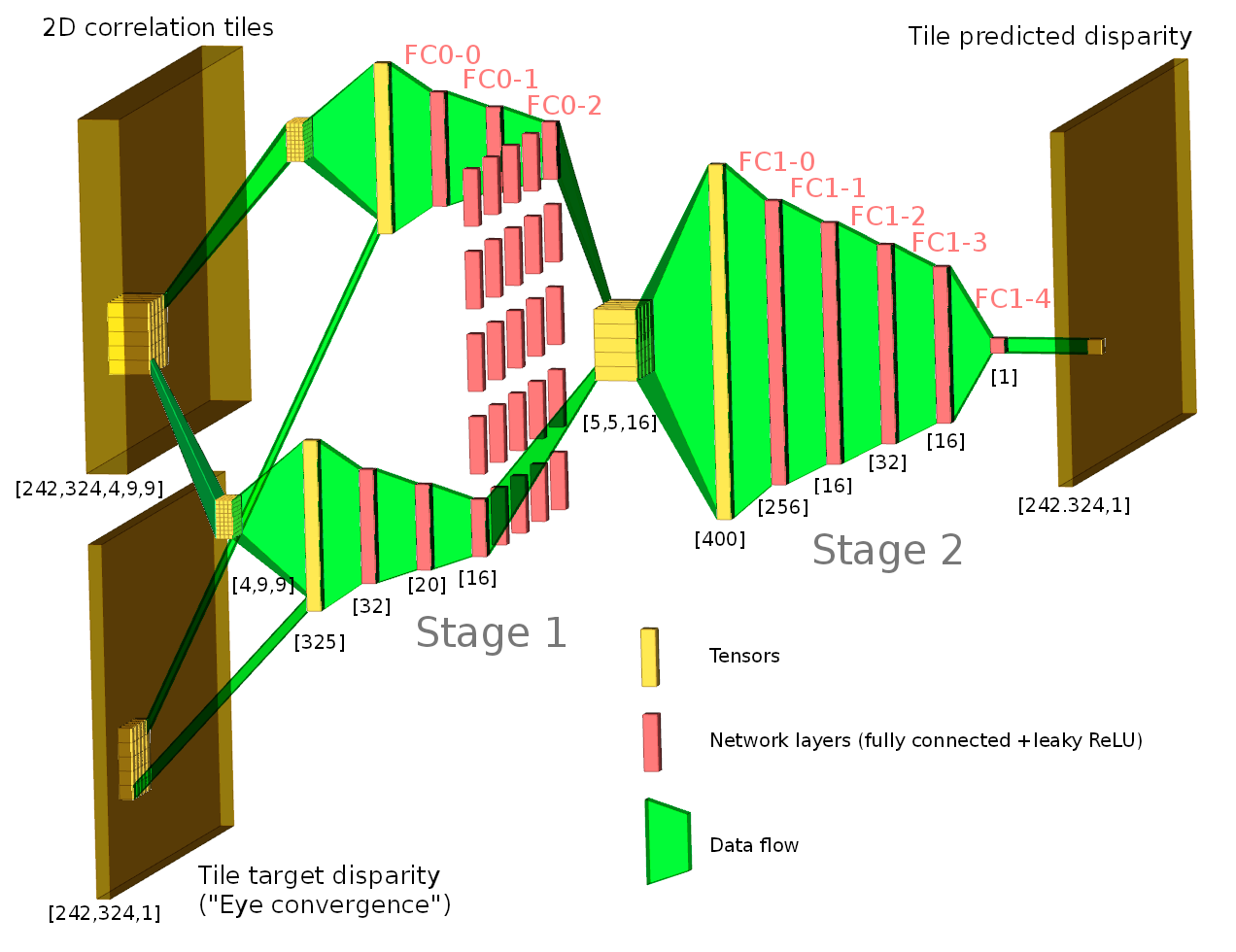}
\end{center}
\caption{TPNET: initial implementation of the network.}
\label{fig:diag-net}
\end{figure}

\subsection{Network part of the TPNET}
\label{subsec:tpnet-net}

The initial implementation of the TPNET is a simple feed-forward connection of
the TP and a 2-stage network shown in Figure~\ref{fig:diag-net}. The main goal
of this implementation was to verify that we can significantly improve the
results achieved by fitting a hand-crafted parametrized model with
Levenberg–Marquardt algorithm (LMA). In the single-pass feed-forward
implementation, each tile disparity is determined by LMA  and then refined by
re-running the tile correlation with updated target disparity until the step
correction falls below the threshold.

The network consists of two stages: the first stage (3-4 fully connected layers
with leaky ReLU activation for all but the last layer) operates on the
correlation data from a single tile; the second convolutional (stride 1) stage
receives the $242\times324\times(16\ldots64)$ tensor (stored in memory)
generated by the first stage. For training, we used it as a 25-head Siamese
network that outputs just a single disparity value for the center tile of
$5\times5$ tiles group; this allowed us to use mining of the input data for the
``hard'' cases (most of the tiles in the scene are almost fronto-parallell, as
shown in Figure~\ref{fig:rmse}). Similar to the Siamese network for stereo
matching by Zbontar and LeCun~\cite{zbontar2015computing}, in the separation
between unary subnets and the second stage that unites them, the intermediate
tensor has to be updated only when the corresponding input data changes. Unlike
other networks, TPNET starts from 2D correlation data of multiple pre-shifted
camera pairs and the stage 2 input features concatenation serves as a partially
generalized 3D data exchange between neighbor tiles to improve disparity
precision for the foreground edges that span multiple tiles and to handle
low-textured areas by consolidating the correlation output from the groups of
the neighboring tiles.

The first stage input features for a tile is a concatenation of the flattened
$9\times9\times4$ correlation tensors (4 epipolar directions for each of the
$9\times9$ symmetrically cropped 2D PC cells) and a single target
disparity (``eye convergence'') value that was used by the TP to pre-shift the
image patches before correlating them. The output disparity is a sum of the
target disparity and the residual disparity calculated from the correlation
data, but the absolute disparity is still an important feature that modifies the
network response to the same PC data.
For the network output, we tried both the full disparity (sum of the
target one and the residual) and just the residual with external summation.
The last variant resulted in faster learning that we attributed to the
observation that the network can still output reasonable differential result
even without the target disparity knowledge.
We used FC layers because
while
the
correlation data is image-like it has
strong translation asymmetry -- the shape of the correlation data is influenced
by the window function and random data in mismatched parts of the patches.
PC of the real world objects is still a smooth function, and we
successfully used a cost function term minimizing the Laplacian of the first
layer weights for regularization.

\iftrue
\begin{figure}[t]
\begin{center}
\includegraphics[width=0.95\linewidth]{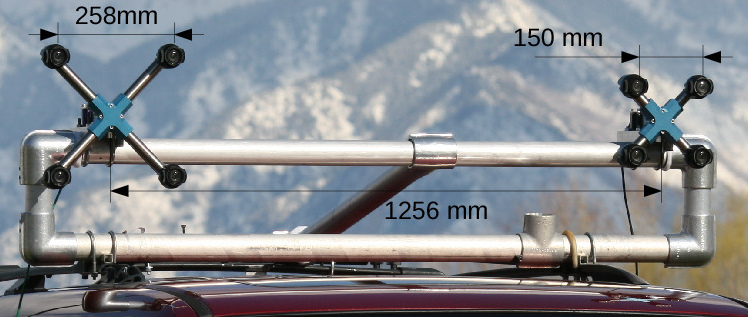}
\end{center}
\caption{Experimental dual-quad stereo rig.}
\label{fig:8rig}
\end{figure}
\fi

\section{Experiments}
\label{sec:experiments}
Most of the trainable networks for stereo 3D perception use LIDAR data as ground
truth (GT) for disparity prediction, including all that are based on the
KITTY~2015~\cite{Menze2015CVPR}, such as recent work of
Smolyanskiy~\etal~\cite{smolyanskiy2018importance} that proves the advantages of
stereo as compared to mono imaging even when combined with \mbox{LIDAR} direct
depth measurements for autonomous vehicles. In our experiments, we were
primarily interested in extremely long range 3D perception, with the
narrow-baseline camera at several hundred to thousands of meters.
The automotive LIDAR range is normally under 200~m, so we used a different
setup, as shown in Figure~\ref{fig:8rig}. We mounted a frame through vibration
isolators carrying a pair of quad cameras at a distance between their centers
4.87 times the tested quad camera baseline of 258~mm.
The disparity accuracy of the composed camera (each of the 8~sensors has the
same resolution and is paired with the identical lens) is expected to be
proportionally higher than that of the single quad camera, and we used the
distance data from the composite camera calculated with traditional software as
GT.
In the examples in Figures~\ref{fig:rmse},~\ref{fig:disparity_map} disparity
values shown for the GT are scaled to match those of the 258~mm baseline camera
for the same real world distances.

Over the course of the experiments, we found that the mechanical stability of
the dual camera rig was lower than that of the individual cameras; consequently,
we had to use field calibration (bundle adjustment of the relative pose)
for each scene. We will improve GT accuracy by using SfM approach and
ignoring moving objects during training.

We had 266 scenes processed and split the dataset in 80\%/20\% for training and
testing. Instead of the full images, we used clusters of $5\times5$~tiles
corresponding to $48\times48$~pixels image patches. The batches were shuffled to
maintain the same representation of different disparity/confidence combinations.
As most of the tile groups in the images belong to the smooth almost
fronto-parallel surfaces,
we performed mining for the rare cases of high disparity
difference and increased the representation of such clusters while
simultaneously lowering their weight in the cost function.
The \textit{weights} graph in Figure~\ref{fig:rmse} illustrates the occurrence
of the tile clusters as a function of the difference between the maximal and the
minimal disparities.

When the same tile contains objects with disparity difference that is too small
to be resolved in the PC output, the maximums merge and the fractional
pixel \textit{argmax} corresponds to nonexistent disparity between foreground
and background objects. Example of this low-pass filter (LPF) effect is visible
in Figure~\ref{fig:disparity_map}(c,d) on the right vertical edge of the
building at $X=163,Y=100\ldots110$. Training with just normalized to tile
occurrence MSE cost for the available number of training samples did not remove
the LPF from the predicted disparity output, so we added an additional term
penalizing for the predicted disparity values between the GT value and that
value mirrored around the average of 8 neighbor tiles. This cost function
tweaking almost completely eliminated LPF effect in
Figure~\ref{fig:disparity_map}(f).

Another cost modification is to improve convergence and delay overfitting.
Stage~2 subnet shown in Figure~\ref{fig:diag-net} has a receptive field of
$5\times5$ Stage 1 outputs. Assuming that even a single center tile should be
sufficient to provide a reasonable disparity estimate, we added two shared
weights clones of Stage~2 - one with all but the center input tile masked out,
and the other with the nine center tiles being non-zero. The same cost function
was applied to all three outputs and the results were mixed with specified
weights. For inference, only the original full Stage~2 subnet was kept.

Most image sets were captured during driving, so the precise matching is
influenced by the rolling shutter (ERS). We have four sub-cameras synchronized
and mechanically aligned to have parallel scan line directions matching (to
$\pm2$ pixels over the sensor width) and vertical WoI settings adjusted, making
ERS caused by the camera egomotion (rotation)  influence on disparity
calculation negligible for objects at infinity (as they are captured
simultaneously in all 4 channels). For the near objects, this effect is still
small for the horizontal pairs, and it is possible to compensate it for the
vertical ones by calculating the correction simultaneously with the 3D scene
reconstruction. This correction is not yet implemented, and in this work we
limited the maximal disparity to 5~pix that corresponds to 100~m range.

\section{Results}
\label{sec:results}
\begin{figure}[t]
\begin{center}
\includegraphics[width=1.00\linewidth]{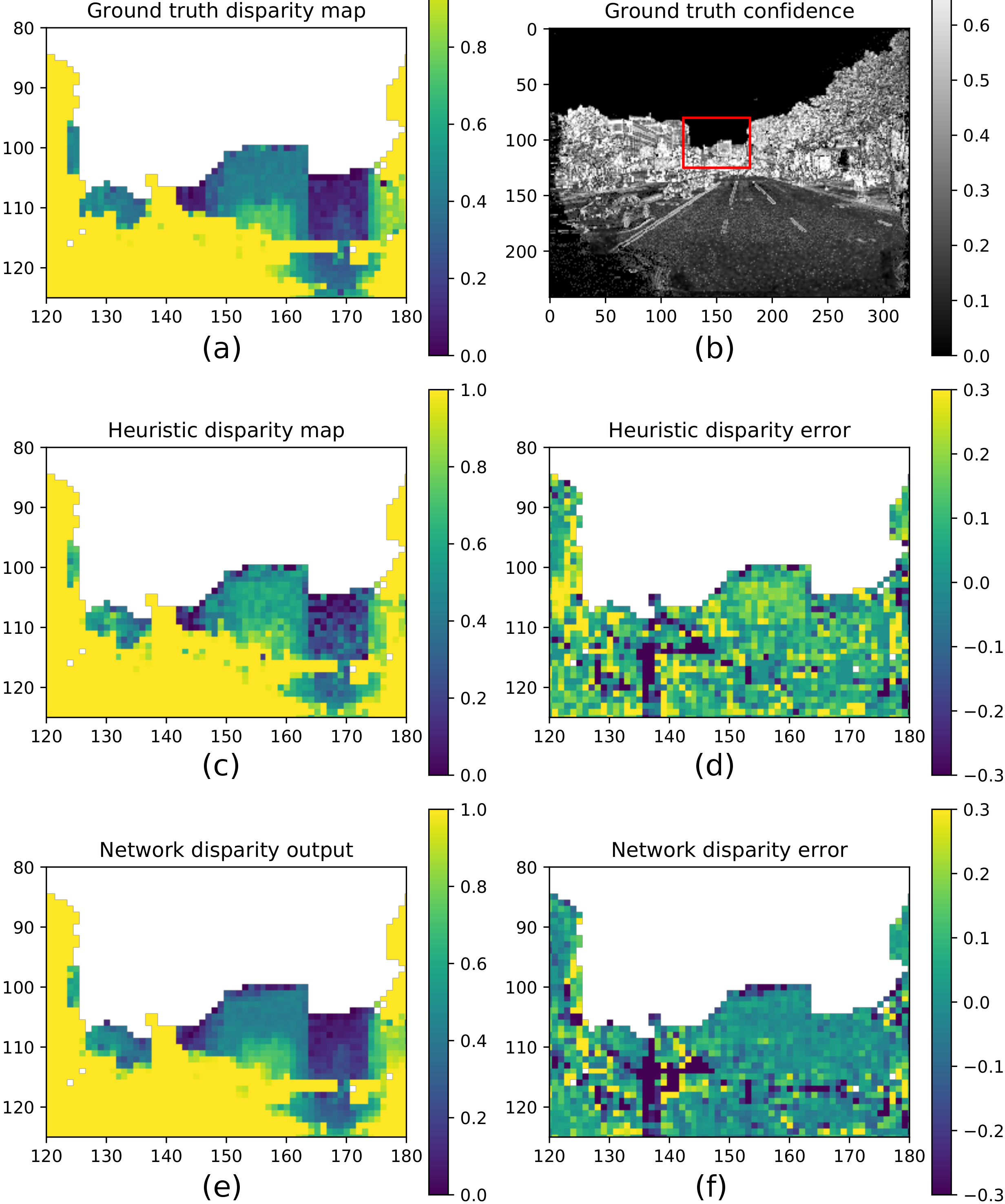}
\end{center}
\caption{Disparity map comparison: TPNET vs heuristics. (a) Ground truth
disparity. (b) Ground truth confidence and region of interest. (c) Absolute
disparity calculated with traditional fitting. (d) Heuristic disparity error,
difference between (c) and (a). (e) Network disparity prediction. (f) Network
disparity error, difference between (e) and (a).}
\label{fig:disparity_map}
\end{figure}

\begin{figure}[t]
\begin{center}
\includegraphics[width=0.95\linewidth]{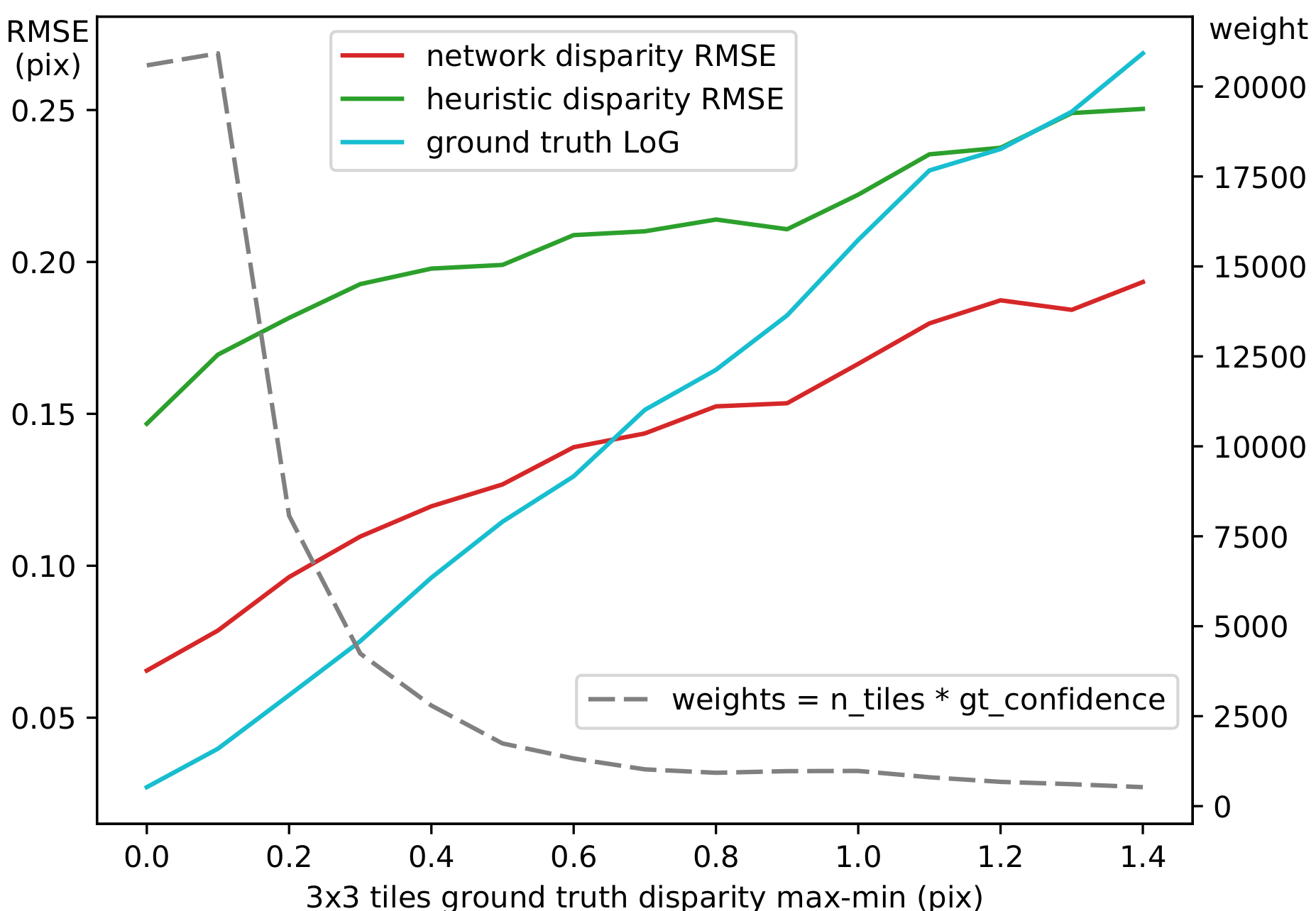}
\end{center}
\caption{Disparity errors dependence on local ground truth disparity
variations.}
\label{fig:rmse}
\end{figure}

Experimental results are presented in Figures~\ref{fig:disparity_map}
and~\ref{fig:rmse}.
Figure~\ref{fig:disparity_map}
shows the GT and the predicted
disparities for the far objects (680\ldots2200~m range) captured by a
forward-looking camera while driving in an urban environment,
\ref{fig:disparity_map}(b) shows the full FoV as GT confidence and
marks the rectangular area enlarged in the other sub-images.
Coordinate ticks designate tile horizontal and vertical indices. The building at
(155,105) has disparity of 0.5~pix (1000~m), the faint one at (170,112) is the
State Capitol at 2200~m from the camera. The full disparity calculated with the
traditional algorithm (our best variant) is shown in \ref{fig:disparity_map}(c),
trained network predictions -- in \ref{fig:disparity_map}(e), and
\ref{fig:disparity_map}(d,f) contain the differences between the predicted
disparities and the GT one.

Figure~\ref{fig:rmse} contains the statistical results accumulated from the
multiple test image sets. The horizontal axis represents the difference between
the maximal and the minimal disparity in the $3\times3$ group. The
\textit{weights} graph illustrates that most tiles are
fronto-parallel and define the overall scene MSE. The RMSE graphs show the
dependence of the prediction errors on the disparity variations around the
tiles; \textit{ground truth LoG} indicates non-flatness -- it is the RMS of the
difference between the tile and the mean of its eight neighbors.

The TPNET for training is implemented with Tensorflow/Python, and the inferred
network is additionally tested with Tensorflow/Java. Processing all tiles in a
frame with GeForce GTX 1050 Ti (compute capability 6.1, 4GB memory) Stage 1
takes 0.46~s, Stage 2 -- 0.12~s run time, with the most time spent on CPU-to-GPU
memory transfers. For the same GPU, the total processing time will be under
0.15~s when the network will be fed from the GPU implementation of the TP,
bypassing large data transfers between the CPU and GPU memories. Separately
tested GPU implementation of the TP took 0.087~s to process four of 5~Mpix
images, including MCLT, aberration correction and IMCLT.

\section{Discussion}
\label{sec:discussion}
Most of the work in the area of ML applications to image processing and 3D
perception is shaped by the available datasets and COTS devices, such as
cellphone cameras. Restricting attention to these datasets limits the diversity
and reach of research in this field. Another problem that we target is poor
interface between the hardware and low-level image processing on one side, and
the advanced networks on the other.
It
pushes researchers to use
full
end-to-end approaches that lead to
significant increases in the required labeled data and cause ``brittleness'' of
the trained networks, requiring re-training for different hardware instances.

We introduce TPNET as an effective interface to partition the vision perception
system into hardware-specific and hardware-invariant modules and prove that this
combination is more efficient than implementing each part separately.

This initial TPNET has multiple limitations, as we were focusing on the new and
untested components and assuming that it will be possible to later add known
functionality, such as fusion of the multi-modal images~\cite{gu2017learning,
park2011high, yang2007spatial}. Another limitation is that the TPNET still
relies on non-ML code for the initial disparity estimation - this task can be
better performed by the trainable networks.
The same
is true for the higher level task of 3D perception and reconstruction of the
final model - it is also not converted from the over-complex heuristics to the
ML. We plan to add feedback from the downstream semantic network to the
TPNET for the fine-tuning of the PC processing.
Segmentation such as described by Miclea and Nedevschi~\cite{miclea2017semantic}
(vegetation, poorly textured pavement, thin wires, vertical poles) may be
applied to TPNET and used to modify the interpretation of the raw 2D correlation
of the tile clusters.

The camera's initial calibration currently depends on the special target
pattern, and the field calibration uses traditional code. It is tempting to try
a GAN-inspired (Goodfellow~\etal~\cite{goodfellow2014generative}) adversarial
game between the hardware-invariant subnet that detects discrepancies in the
real world representation and the calibration network that adjusts the
hardware-dependent parameters to avoid that detection.

\section{Acknowledgments}

\label{sec:acknowledgments}
We thank Tolga Tasdizen for his his suggestions on the network
architecture and implementation.

{\small
\bibliographystyle{ieee}
\bibliography{tpnet}
}

\end{document}